\documentclass[11pt,a4paper]{article}
\usepackage[hyperref]{emnlp2018}
\usepackage{times}
\usepackage{latexsym}

\usepackage{url}
\usepackage{graphicx}
\graphicspath{ {images/} }

\usepackage{booktabs}

\aclfinalcopy 

\title{Similarity measure for Persons with time parameter}

\author{Andreas St\"ockl \\
{\tt andreas.stoeckl@fh-hagenberg.at} \\
       School of Informatics, Communications and Media\\
       University of Applied Sciences Upper Austria\\
       Softwarepark 11, 4232 Hagenberg, Austria  \\
}

\date{}

\begin{document}

\title{Similarity measure for Public Persons}

\maketitle

\begin{abstract}
For the webportal ``Who is in the News!'' with statistics about the appearence of persons in written news we developed an extension, which measures
the relationship of public persons depending on a time parameter, as the relationship may vary over time.

On a training corpus of English and German news articles we built a measure by extracting the person's occurrence in the text
via pre-trained named entity extraction and then construct time series of counts for each person. Pearson correlation over a sliding window
is then used to measure the relation of two persons.

\end{abstract}

\section{Motivation}
``Who is in the News!'' \footnote{\label{foot:inthenews} http://in-the-news.stoeckl.ai/} is a webportal with statistics and plots about the appearence of persons in written news articles.
It counts how often public persons are mentioned in news articles and can be used for research or journalistic purposes. 
The application is indexing articles published by ``Reuters'' agency on their website  \footnote{\label{foot:reuters} http://www.reuters.com/}. With the interactive charts users can analyze
different timespans for the mentiones of public people and look for patterns in the data.
The portal is bulit with the Python microframework ``Dash"  \footnote{\label{foot:dash} https://dash.plot.ly/} which uses the plattform 
``Plotly"  \footnote{\label{foot:plotly} https://plot.ly/} for the interactive charts.

\begin{figure}
	\centering
	\includegraphics{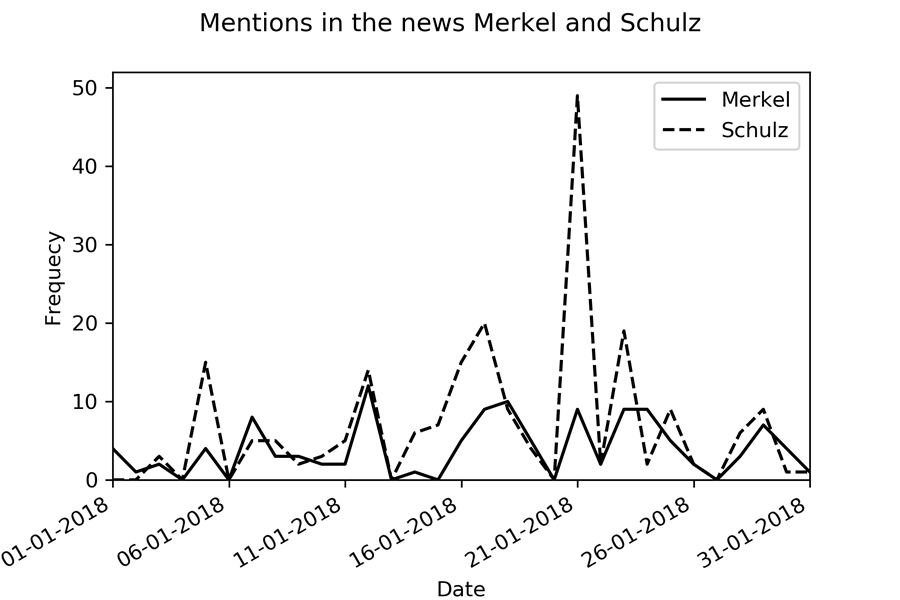}
	\caption{Mentions of Merkel and Schulz in 1/2018}
	\label{img:merkelschulz}
\end{figure}

Playing around with the charts shows some interresting patterns like the one in the example of Figure \ref{img:merkelschulz}. This figure suggests that there must be some relationship
between this two persons. In this example it is obvious because the persons are both german politicians and candidates for the elections.

This motivated us to look for suitalbe measures to caputure how persons are related to each other, which then can be used to exted the webportal with charts 
showing the person to person relationships. Relationship and distance between persons have been analyzed for decades, for example \cite{travers1967small} looked at distance in the famous experimental study 
``the Small World Problem''. They inspected the graph of relationships between different persons and set the ``distance'' to the shortest path between them.

Other approaches used large text corpora for trying to find connections and relatedness by making statistics over the words in the texts. This of course only works for people appearing 
in the texts and we will discuss this in section \ref{Relatedwork}. All these methods do not cover the changes of relations of the persons over time, that may change over the years. Therefore the measure should have a time parameter, which can be set to the desired time we are investigating.

\begin{figure}
	\centering
	\includegraphics{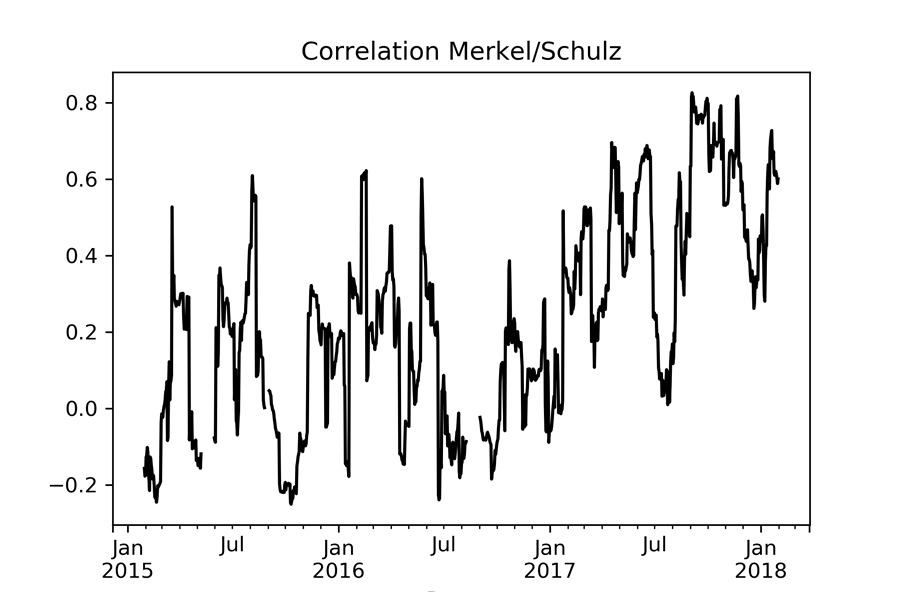}
	\caption{Correlation for Merkel and Schulz}
	\label{img:overtime}
\end{figure}

We have developed a method for such a  measure and tested it on a set of news articles for the United States and Germany.
In Figure \ref{img:overtime} you see how the relation changes in an example of the German chancellor ''Angela Merkel''
and her opponent on the last elections ``Martin Schulz''. It starts around 0 in 2015 and goes up to about 0.75 in 2017 as we can expect looking at the high correlated time series chart
in Figure \ref{img:merkelschulz} from the end of 2017.

\section{Related work} \label{Relatedwork}

\begin{figure}
	\centering
	\includegraphics{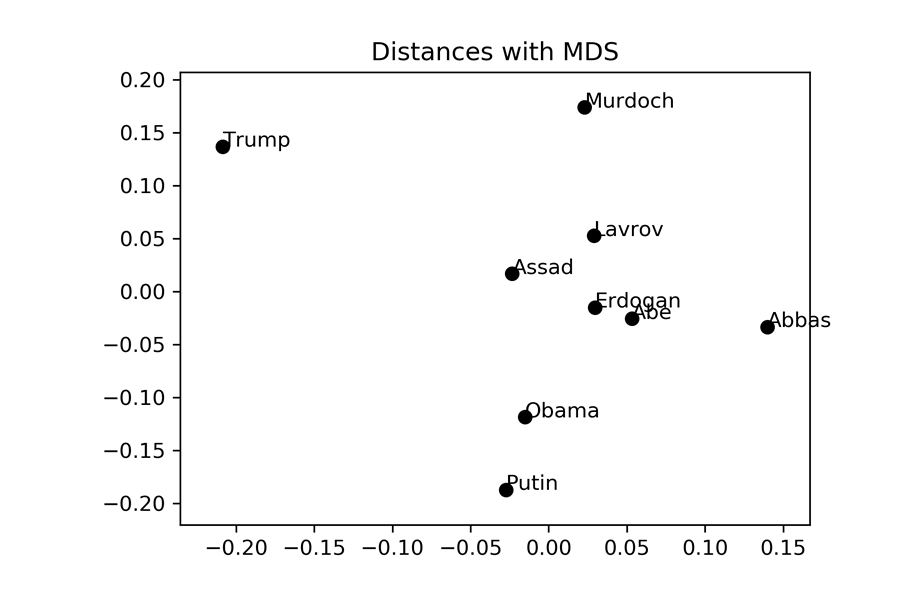}
	\caption{Distances with MDS}
	\label{img:person_mds}
\end{figure}

There are several methods which represent words as vectors of numbers and try to group the vectors of similar words together in vector space. Figure \ref{img:person_mds}
shows a picture which represents such a high dimensional space in 2D via multidimensional scaling \cite{borg2005modern}. The implementation was done with 
Scikit Learn \footnote{http://scikit-learn.org/} \cite{pedregosa2011scikit, geron2017hands, raschka2017python}.
Word vectors are the building blocks for a lot of applications in areas like search, sentiment analysis and recommendation systems.

The similarity and therefore the distance between words is calculated via the cosine similarity of the associated vectors, which gives a number between -1 and 1.
The word2vec tool \footnote{https://code.google.com/archive/p/word2vec/} was implemented by  \cite{mikolov2013distributed, mikolov2013efficient, mikolov2013linguistic} and trained over a Google News dataset with about 100 billion words. They use global matrix factorization or local context window methods
for the training of the vectors.

A trained dictionary for more than 3 million words and phrases with 300-dim vectors is provided  for download. We used the Python library Gensim \footnote{https://radimrehurek.com/gensim/} from \cite{rehurek2011gensim} for the calculation of the word distances of the multidimensional scaling in Figure \ref{img:person_mds}.

\cite{pennington2014glove} combine the global matrix factorization and local context window methods in the "GloVe" method for word representation \footnote{https://nlp.stanford.edu/projects/glove/}.

\cite{hasegawa2004discovering} worked on a corpus of newspaper articles and developed a method for unsupervised relation discovery between named entities of different types
by looking at the words between each pair of named etities. By measuring the similarity of this context words they can also discover the type of relatoionship. For example a person entity and
an organization entity can have the relationship ``is member of''. For our application this interesting method can not be used because we need additional time information.

\cite{zelenko2003kernel} developed models for supervised learning with kernel methods and support vector machines for relation extraction and tested them on problems
of person-affiliation and organization-location relations, but also without time parameter.

\section{Dataset and Data Collection} \label{Dataset}

We collected datasets of news articles in English and German language from the news agency Reuters (Table \ref{table:1}).
After a data cleaning step, which was deleting meta information like author and editor name from the article, title, body and date
were stored in a local database and imported to a Pandas\footnote{\label{foot:pd} https://pandas.pydata.org/} data frame \cite{mckinney2012python}.
The English corpus has a dictionary of length 106.848, the German version has a dictionary of  length 163.788.

\begin{table}
\small
\centering
\begin{tabular}{|l|l|r|}
\hline \bf URL & \bf Date &  \bf No. Articles \\ \hline
de.reuters.com & 2015 to 2018 & 34058 \\
www.reuters.com &  2016 to 2018 & 36229\\
\hline
\end{tabular}
\caption{News articles}
\label{table:1}
\end{table}

For each article we extracted with the Python library ``Spacy'' \footnote{\label{foot:spacy}  https://spacy.io} the named entities labeled as person.
``Spacy'' was used because of its good performance \cite{jiang2016evaluating} and it has pre-trained language models for English, German and others.
The entity recognition is not perfect, so we have errors in the lists of persons. In a post processing step the terms from a list of common errors are removed.
The names of the persons appear in different versions like ``Donald Trump'' or ``Trump''. We map all names to the shorter version i.e. ``Trump'' in this example.

\begin{figure}
	\centering
	\includegraphics{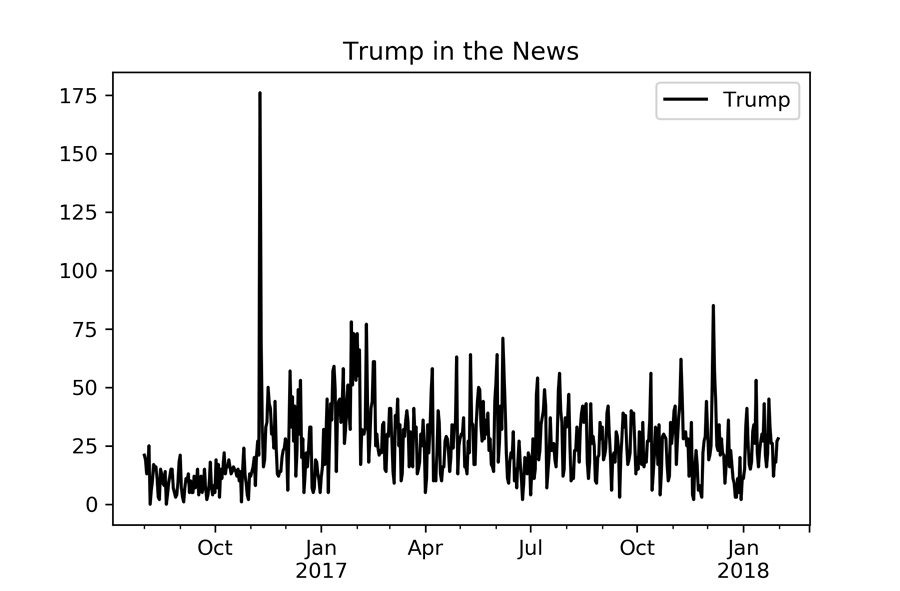}
	\caption{Mentions of Trump}
	\label{img:trump}
\end{figure}

In Figure \ref{img:trump} you can see the time series of the mentions of ``Trump'' in the news, with a peak at the 8th of November 2016 the day of the election.
It is also visible that the general level is changing with the election and is on higher level since then.

\begin{figure}
	\centering
	\includegraphics{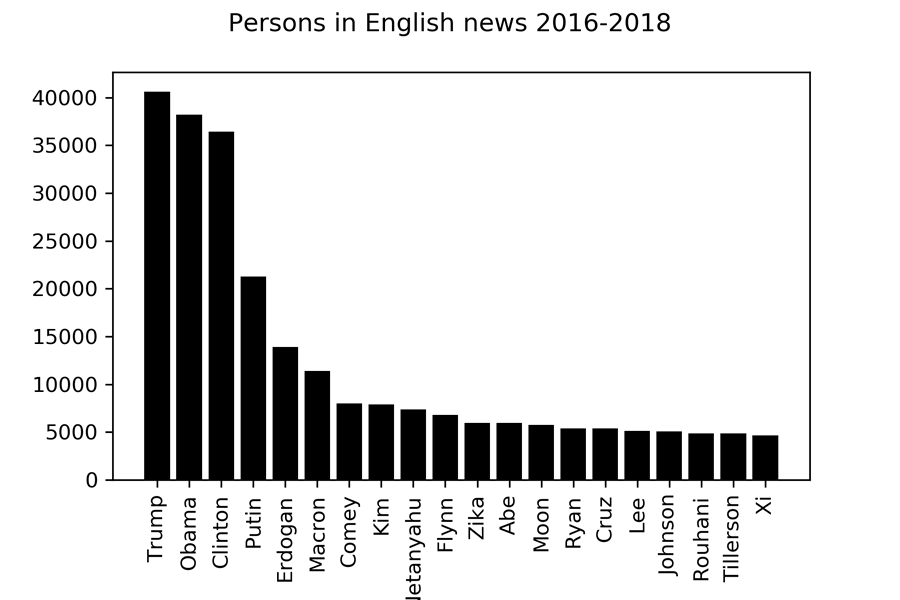}
	\caption{Histogram of mentions in the news}
	\label{img:personnews}
\end{figure}

Taking a look at the histograms of the most frequent persons in some timespan shows the top 20 persons in the English news articles
from 2016 to 2018 (Figure \ref{img:personnews}).
As expected the histogram has a distribution that follows Zipfs law \cite{adamic2002zipf, li2002zipf}.

From the corpus data a dictionary is built, where for each person the number of mentions of this person in the news  per day is recorded. This time series data
can be used to build a model that covers time as parameter for the relationship to other persons.

\section{Building the Model}  \label{Model}

Figure \ref{img:obamatrump} shows that the mentions of a person and the correlation with the mentions of another person varies
over time. We want to capture this in our relation measure. So we take a time window of $n$ days and look at the time series in the segment back in time as shown in the example of Figure \ref{img:merkelschulz}.

For this vectors of $n$ numbers for persons we can use different similarity measures. This choice has of course an impact of the results in applications
\cite{strehl2000impact}. A first choice could be the cosine similarity as used in the word2vec implementations \cite{mikolov2013distributed}.
We propose a different calculation for our setup, because we want to capture the high correlation of the series even if they are on different absolute levels of 
the total number of mentions, as in the example of Figure \ref{img:obamatrump_zoom}. 

We propose to use the Pearson correlation coefficient instead.
We can shift the window of calculation over time and therefore get the measure of relatedness as a function of time.

\section{Results}  \label{results}

\begin{table*}[h!]
\centering

\begin{tabular}{lrrrrrrrrr}
\toprule
{} &  Abbas &   Abe &  Assad &  Erdogan &  Lavrov &  Murdoch &  Obama &  Putin &  Trump \\
Name    &        &       &        &          &         &          &        &        &        \\
\midrule
Abbas   &   1.00 & -0.20 &  -0.04 &     0.22 &    0.21 &     0.07 &   0.24 &   0.20 &   0.80 \\
Abe     &  -0.20 &  1.00 &   0.27 &    -0.15 &   -0.12 &     0.60 &  -0.14 &   0.48 &  -0.04 \\
Assad   &  -0.04 &  0.27 &   1.00 &     0.05 &   -0.03 &     0.26 &   0.07 &   0.24 &   0.09 \\
Erdogan &   0.22 & -0.15 &   0.05 &     1.00 &    0.07 &    -0.02 &   0.37 &  -0.25 &   0.28 \\
Lavrov  &   0.21 & -0.12 &  -0.03 &     0.07 &    1.00 &    -0.04 &   0.31 &   0.17 &   0.31 \\
Murdoch &   0.07 &  0.60 &   0.26 &    -0.02 &   -0.04 &     1.00 &  -0.10 &   0.80 &   0.19 \\
Obama   &   0.24 & -0.14 &   0.07 &     0.37 &    0.31 &    -0.10 &   1.00 &  -0.16 &   0.37 \\
Putin   &   0.20 &  0.48 &   0.24 &    -0.25 &    0.17 &     0.80 &  -0.16 &   1.00 &   0.36 \\
Trump   &   0.80 & -0.04 &   0.09 &     0.28 &    0.31 &     0.19 &   0.37 &   0.36 &   1.00 \\
\bottomrule
\end{tabular}

\caption{Similarities of Persons in Dec. 2017}
\label{table:2}
\end{table*}

Figure \ref{img:overtime} shows a chart of the Pearson correlation coefficient computed over a sliding window of 30 days from 2015-01-01 to 2018-02-26 for the 
persons ``Merkel'' and ``Schulz''. The measure clearly covers the change in their relationship during this time period. We propose that 30 days is a good value for
the time window, because on one hand it is large enough to have sufficient data for the calculation of the correlation, on the other hand it is sensitive enough to reflect
changes over time. But the optimal value depends on the application for which the measure is used.

\begin{figure}
	\centering
	\includegraphics{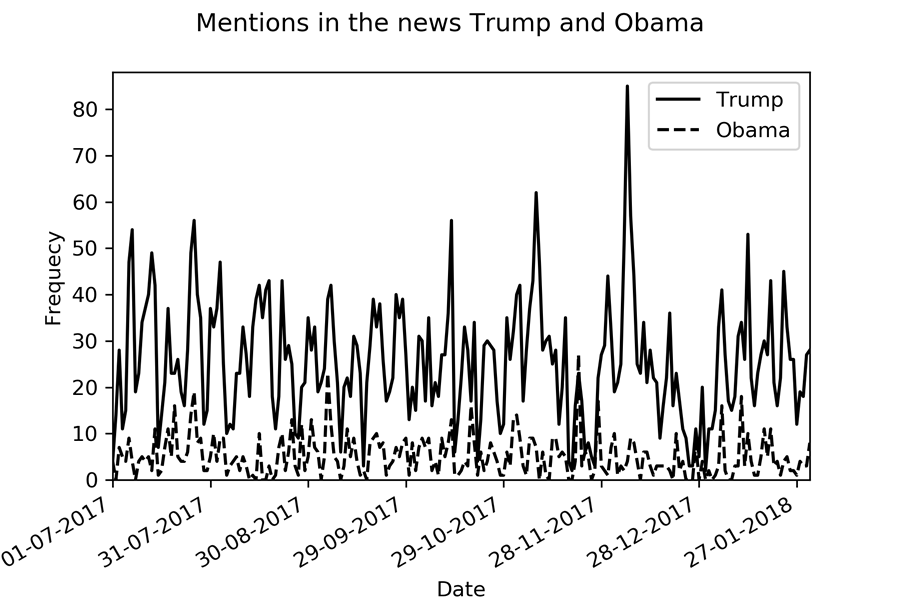}
	\caption{Mentions of Trump and Obama}
	\label{img:obamatrump}
\end{figure}

\begin{figure}
	\centering
	\includegraphics{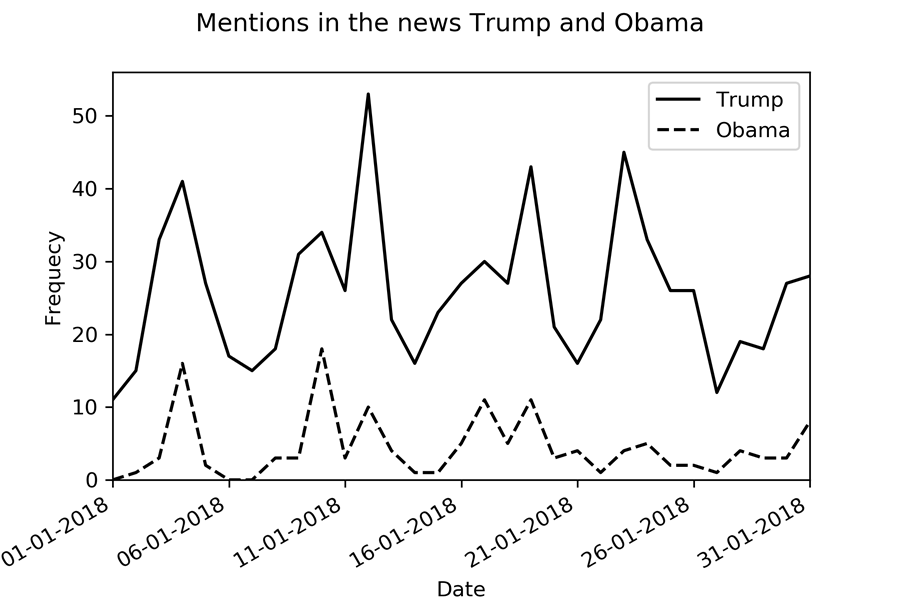}
	\caption{Mentions of Trump and Obama  in 1/2018} 
	\label{img:obamatrump_zoom}
\end{figure}

An example from the US news corpus shows the time series of ``Trump'' and ``Obama'' in Figure \ref{img:obamatrump} and a zoom in to the first month of 2018
in Figure \ref{img:obamatrump_zoom}. It shows that a high correlation can be on different absolute levels. Therefore we used Pearson correlation
to calculate the relation of two persons. 
You can find examples of the similarities of some test persons from December 2017 in Table \ref{table:2}

\begin{figure}
	\centering
	\includegraphics{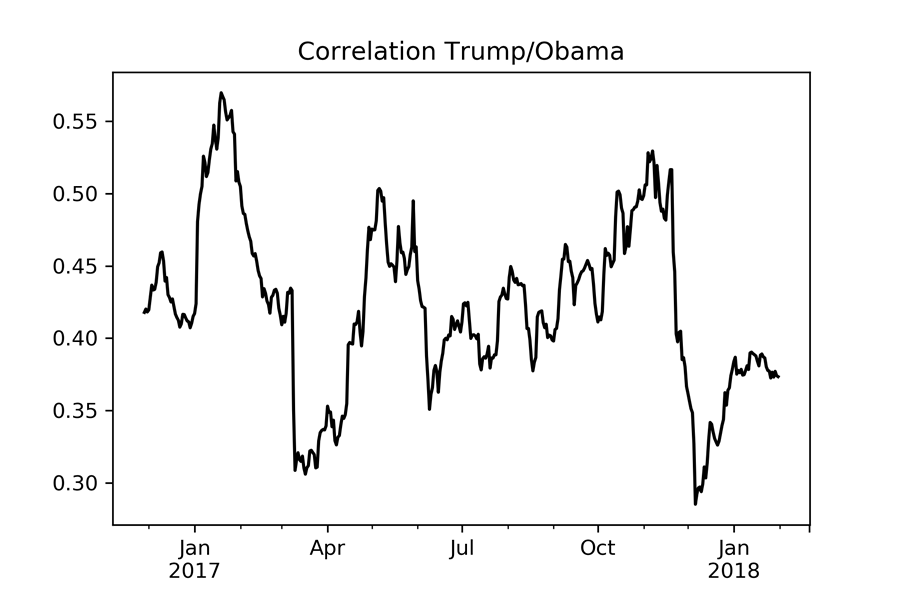}
	\caption{Correlation for Trump and Obama}
	\label{img:trumpovertime}
\end{figure}

The time series of the correlations looks quite ``noisy'' as you can see in Figure \ref{img:overtime}, because the series of the mentions has a high variance. To reflect
the change of the relation of the persons in a more stable way, you can take a higher value for the size of the calculation window of the correlation between the two series. 
In the example of Figure \ref{img:trumpovertime} we used a calculation window of 120 days instead of 30 days.

\section{Future Work}

It would be interesting to test the ideas with a larger corpus of news articles for example the Google News articles used in the word2vec implementation 
\cite{mikolov2013distributed}.

The method can be used for other named entities such as organizations or cities but we expect not as much variation over time periods as with persons.
And similarities between different types of entities would we interesting. So as the relation of a person to a city may chance over time.

\bibliography{persons}
\bibliographystyle{acl_natbib_nourl}

\appendix

\end{document}